\documentclass[runningheads, orivec]{llncs}
\usepackage[T1]{fontenc}
\usepackage{graphicx}
\usepackage{color}
\usepackage{url}
\usepackage{xcolor}

\usepackage{amsmath}
\usepackage{hyperref}
\usepackage{amsfonts}
\usepackage{amssymb}
\usepackage{booktabs}
\usepackage{multirow}
\usepackage{makecell}
\usepackage{threeparttable}
\usepackage[misc]{ifsym}
\begin{document}
\title{Test-Time Adaptation of Spiking Neural Networks for Intracortical Neural Decoding using Membrane Potential Alignment}
\titlerunning{TTA of SNN for Intracortical Neural Decoding}
\author{Guangzhi Tang \Letter}
% % Second Author\inst{2,3}\orcidID{1111-2222-3333-4444} 
% % Third Author\inst{3}\orcidID{2222--3333-4444-5555}}
% % %
% % \authorrunning{F. Author et al.}
% % % First names are abbreviated in the running head.
% % % If there are more than two authors, 'et al.' is used.
% % %
\institute{Department of Advanced Computing Sciences, Maastricht University \\ Maastricht, The Netherlands\\
% % , Princeton NJ 08544, USA \and
% % Springer Heidelberg, Tiergartenstr. 17, 69121 Heidelberg, Germany
\email{guangzhi.tang@maastrichtuniversity.nl}}
\maketitle              % typeset the header of the contribution
\begin{abstract}
Intracortical brain–computer interfaces suffer from day-to-day neural signal shifts that degrade pretrained decoders. Existing unsupervised adaptation methods rely on deep recurrent or adversarial architectures that are too computationally expensive for implantable hardware. We propose Membrane Potential Alignment (MPA), a test-time adaptation method for spiking neural networks that realigns a pretrained decoder to shifted recordings by only matching membrane potential distributions via KL divergence. By restricting updates to low-rank (LoRA) weights, MPA adapts fewer than 9\% of parameters. On a non-human primate reaching task spanning over one month, MPA achieves performance competitive with the state-of-the-art NoMAD method, while using a simpler architecture and finer temporal resolution (4 ms vs.\ 20 ms). These results show that efficient SNN-based test-time adaptation is a practical path toward long-term, recalibration-free brain–computer interfaces.

\end{abstract}

\section{Introduction}

Intracortical brain–computer interfaces (iBCIs) have become a critical technology for restoring communication and motor function in individuals with severe neurological disabilities~\cite{wairagkar2025instantaneous,willsey2025high}. Yet achieving reliable decoding in practice remains challenging, as decoders must handle low signal-to-noise ratios, high-dimensional neural variability, and strict real-time low-power constraints~\cite{patrick2025state}. Once a decoder is trained and the device is chronically implanted, multiple factors can drive domain shifts between training and incoming data. For example, micromotion and hardware-related disruptions alter the recorded units \cite{dunlap2020classifying}, firing rates and spike amplitudes fluctuate over time \cite{perge2013intra}, and progressive glial encapsulation reshapes the tissue–electrode interface \cite{salatino2017glial}. These combined signal changes cause decoder performance to drop over time, yet standard recalibration is impractical for daily use as it requires the user to stop and repeat structured tasks, often with help from a technician to support supervised retraining~\cite{jarosiewicz2015virtual}.

Test-time adaptation (TTA) offers a promising way to address this challenge by updating a pretrained decoder at inference time using only unlabeled neural data, relying on unsupervised objectives such as distribution alignment or entropy minimization to close the gap between training and deployment conditions~\cite{liang2025comprehensive}. This idea has recently been applied to iBCI. For instance, NoMAD~\cite{karpowicz2025stabilizing} first trains a variational autoencoder-based recurrent neural network (RNN) on an initial recording session to capture the latent dynamics of neural activities. At test time, it introduces a feedforward alignment network that maps each new day's spiking data back onto the original manifold. This is achieved by minimizing the KL divergence between latent state distributions alongside a self-supervised reconstruction loss, all without behavioral labels. The approach maintained stable decoding for over three months in non-human primate experiments. However, NoMAD, and other similar approaches~\cite{ma2023using} all rely on complex architectures, including gated RNNs and adversarial networks, that are computationally demanding, making them difficult to deploy on the power- and latency-constrained hardware typical of implantable iBCI systems.

Spiking neural networks (SNNs) offer a compelling alternative for neural decoding, as their event-driven, spike-based computation is inherently sparse and energy-efficient, well matched to the ultra-low power budgets of implantable devices~\cite{rivelli2025adaptively}. Recent work has shown that SNN decoders can match or exceed the accuracy of RNNs on intracortical decoding tasks while consuming orders-of-magnitude less energy~\cite{hueber2024benchmarking}. However, existing TTA methods designed for SNNs have focused almost exclusively on image classification benchmarks, where inputs are static~\cite{luo2025space}. These approaches do not account for the non-stationary nature of neural recordings, leaving a critical gap between the efficiency advantages of SNNs and the adaptation capabilities needed for long-term iBCI deployment.

In this paper, we propose Membrane Potential Alignment (MPA), a test-time adaptation method that aligns a pretrained SNN to shifted neural recordings by matching membrane potential distributions across sessions using KL divergence\footnote{https://github.com/ERNIS-LAB/snn-mpa-intracortical-neural-decoding-icann}. By restricting adaptation to low-rank (LoRA) weights in the first hidden layer, MPA updates fewer than 9\% of model parameters while keeping the pretrained decoder frozen. We benchmark MPA on a non-human primate reaching task spanning over one month and show that it achieves competitive decoding performance with the state-of-the-art NoMAD method, while using a substantially simpler architecture and operating at finer temporal resolution (4 ms vs. 20 ms). With only supervised pretraining on a single day-0 source session, MPA restores decoding on target sessions more than one month later using as few as a few seconds of data. The paper's code will be open-sourced after acceptance.

\section{Methodology}

\subsection{Test-Time Adaptation for Non-Human Primate Reaching}

\begin{figure}[t]
    \centering
    \includegraphics[width=1\linewidth]{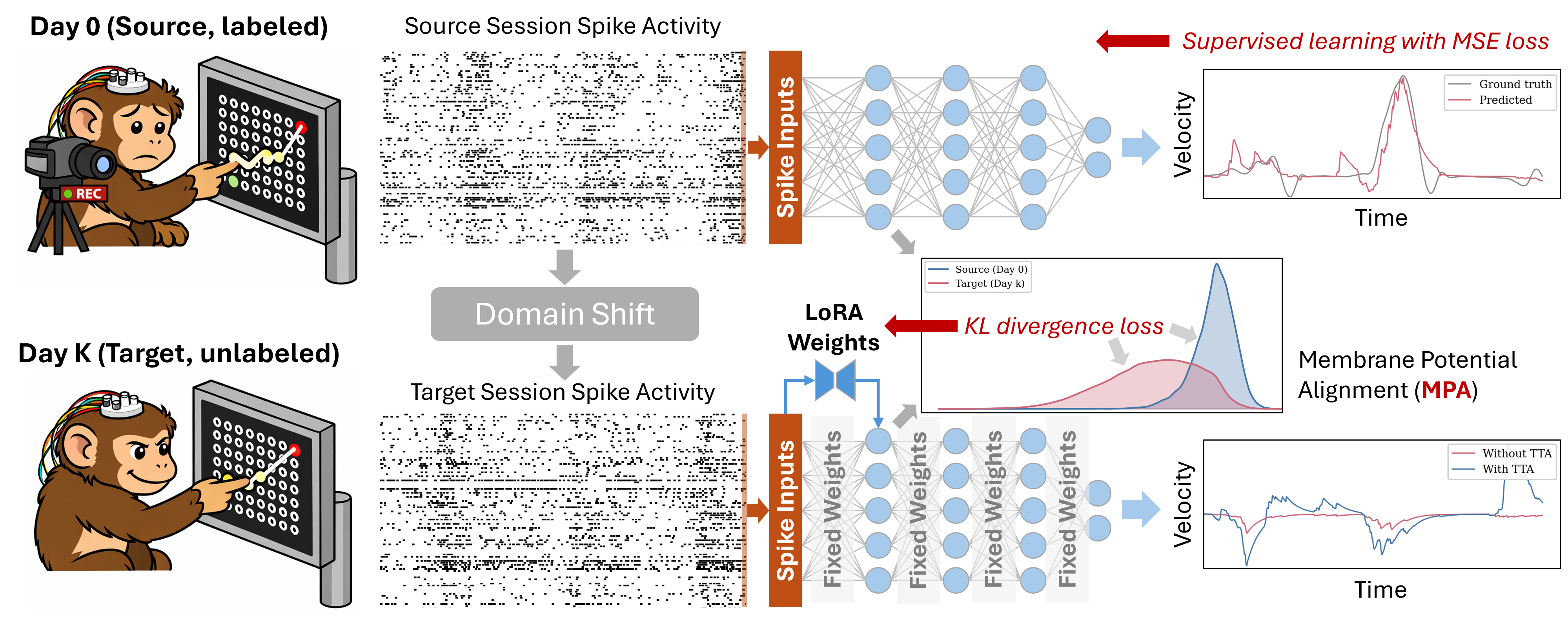}
    \caption{Overview of the proposed cross-day SNN test-time adaptation pipeline. An SNN pretrained on day 0 is applied to a later target session, day k, where cross-session drift causes a mismatch in neural inputs and labels are unavailable. We therefore keep the decoder fixed and adapt the SNN's LoRA weights by aligning source and target membrane potential distributions through membrane potential alignment (MPA), improving target-session decoding without supervised behavior labels.}
    \label{fig:method}
\end{figure}

We propose a TTA method for unsupervised SNN adaptation on a target session using a model pretrained on the day-0 source session, as illustrated in Figure~\ref{fig:method}. During day-0 pretraining, the model is trained in a supervised manner with ground-truth targets, allowing the SNN to learn the mapping from neural activity to the intended motor output. This produces a decoder that performs well on the source session, but its performance can degrade on later days because neural activity statistics change across sessions. In a realistic deployment setting at day k on the same subject, target labels are not available; instead, the model only has access to unlabeled spike recordings and the corresponding SNN internal dynamics from the source and target sessions. The goal of day-k alignment is therefore not to retrain the decoder from scratch, but to adapt it to the shifted input distribution while preserving the decoding function learned on day 0. To achieve this, we align the membrane potential distributions between source and target sessions using the MPA method proposed in this work, which will be described in detail in Section~\ref{sec:mpa}.

\subsection{Spiking Neural Network for Behavior Decoding}

We adapt SNN training in~\cite{hueber2024benchmarking} for day-0 pretraining. The SNN processes binary spike trains from the neural recording and predicts continuous hand velocities at each timestep. We use the leaky integrate-and-fire (LIF) neuron model for the SNN. For a neuron $i$ in the network, the membrane potential $u_i$ evolves as:
\begin{equation}
    u_i(t) = \beta \, u_i(t-1) + I_i(t) - S_i(t-1) V_{\mathrm{th}}
    \label{eq:lif}
\end{equation}
where $\beta$ is the membrane decay, $I_i(t)$ is the input current at timestep $t$, $V_{\mathrm{th}}$ is the firing threshold, and $S_i(t-1)$ is the spike output at the previous timestep.

Unlike the hard reset used for LIF neurons in~\cite{hueber2024benchmarking}, we use the soft reset that preserves the residual membrane potential after a spike. This produces more continuous membrane potential distributions, which provide more informative KL-divergence gradients during MPA training for target day k sessions. In contrast, a hard reset forces all suprathreshold states back to zero, making the distribution less expressive and reducing the sensitivity of the KL objective to distribution shift.

\subsection{Membrane Potential Alignment using LoRA Weights}
\label{sec:mpa}

Let $f_{\boldsymbol{\theta}}$ denote the pretrained SNN decoder with parameters $\boldsymbol{\theta}$, trained on day 0 source session. When applied to day k target session, the distribution shift in the input spike patterns propagates through the network, altering the statistics of the internal membrane potentials. We seek an adapted model $f_{\boldsymbol{\theta}+\Delta\boldsymbol{\theta}}$ using MPA such that the membrane potential distributions of the SNN on day k match those of the original model on day 0, thereby restoring the learned input-output mapping without requiring new labels.

The key insight of our MPA approach is that the membrane potentials of spiking neurons provide a rich, continuous-valued representation of the network's internal state that is well-suited for distribution matching. For a given hidden layer $l$, let $\mathbf{u}^{(l)}(t) \in \mathbb{R}^{H}$ denote the vector of membrane potentials at timestep $t$. Over a batch of $B$ input samples with timestep $T$, we collect the membrane potential vectors across all samples and timesteps, yielding a set of $N = B \times T$ observations $\{\mathbf{u}^{(l)}_n\}_{n=1}^{N}$ in $\mathbb{R}^{H}$. We model the membrane potential distribution as a multivariate Gaussian:
\begin{equation}
    p(\mathbf{u}^{(l)}) = \mathcal{N}\!\left(\boldsymbol{\mu}^{(l)},\, \boldsymbol{\Sigma}^{(l)}\right)
    \label{eq:gaussian}
\end{equation}
where the mean and covariance are estimated from the collected samples:
\begin{equation}
    \boldsymbol{\mu}^{(l)} = \frac{1}{N} \sum_{n=1}^{N} \mathbf{u}^{(l)}_n, \qquad
    \boldsymbol{\Sigma}^{(l)} = \frac{1}{N-1} \sum_{n=1}^{N} \left(\mathbf{u}^{(l)}_n - \boldsymbol{\mu}^{(l)}\right)\!\left(\mathbf{u}^{(l)}_n - \boldsymbol{\mu}^{(l)}\right)^\top
    \label{eq:stats}
\end{equation}

To align the adapted model's internal representations with the pretrained model's, we minimize the Kullback--Leibler (KL) divergence between the day 0 reference distribution and the day k adapted distribution. Let $p_0^{(l)} = \mathcal{N}(\boldsymbol{\mu}_0^{(l)}, \boldsymbol{\Sigma}_0^{(l)})$ denote the membrane potential distribution obtained by running the frozen pretrained model on day~0 data, and $p_k^{(l)} = \mathcal{N}(\boldsymbol{\mu}_k^{(l)}, \boldsymbol{\Sigma}_k^{(l)})$ denote the distribution obtained by running the adapted model on day k. The MPA adaptation loss is:
\begin{equation}
    \mathcal{L}_\mathrm{MPA} = D_\mathrm{KL}\!\left(p_0^{(l)} \,\|\, p_k^{(l)}\right)
    \label{eq:tta_loss}
\end{equation}

Directly fine-tuning all hidden weights with MPA risks overfitting to the limited unlabeled adaptation data and catastrophically forgetting the pretrained representations. We instead adopt Low-Rank Adaptation (LoRA)~\cite{hu2022lora}, which parameterizes the weight update as a low-rank matrix. For a pretrained weight matrix $\mathbf{W} \in \mathbb{R}^{d_\mathrm{out} \times d_\mathrm{in}}$, the adapted weight is:
\begin{equation}
    \mathbf{W}' = \mathbf{W} + \mathbf{A}\mathbf{B}
    \label{eq:lora}
\end{equation}
where $\mathbf{A} \in \mathbb{R}^{d_\mathrm{out} \times r}$ and $\mathbf{B} \in \mathbb{R}^{r \times d_\mathrm{in}}$ are the trainable low-rank matrices with rank $r \ll \min(d_\mathrm{in}, d_\mathrm{out})$, and the original weight $\mathbf{W}$ remains frozen. Matrix $\mathbf{A}$ is initialized to zero so that $\mathbf{A}\mathbf{B} = \mathbf{0}$ at initialization, ensuring the adapted model starts with identical behavior to the pretrained model. This reduces the number of trainable parameters from $d_\mathrm{out} \times d_\mathrm{in}$ to $r \times (d_\mathrm{out} + d_\mathrm{in})$, providing an effective regularization that constrains the adaptation to a low-dimensional subspace.

As shown in Figure~\ref{fig:method}, we apply LoRA to the first hidden layer of the network. Since the first layer directly receives the neural spike input, it is most affected by the domain shifts. The remaining layers, which learn more abstract behavior-level representations, are kept frozen to preserve the pretrained decoding capability. This yields a highly parameter-efficient adaptation, where only a small fraction of the total model parameters are updated.

\subsection{Source-Target Session Compatibility}

TTA assumes that the source model trained on the day-0 session captures enough of the underlying neural structure to serve as a useful starting point for the target session. However, if the recording array has shifted or the neural population has reorganized, TTA can fail silently, producing confident but incorrect velocity predictions. We therefore need a way to assess, before adaptation, whether a given day-0 (source) session is a reasonable match for the day-k (target) session. We adopt two complementary metrics for this purpose: representational similarity~\cite{diedrichsen2017representational}, which checks whether channel-level relationships are preserved, and Grassmann distance~\cite{hamm2008grassmann}, which measures how much the low-dimensional neural subspace has rotated between sessions.

To capture whether the pairwise statistical relationships among channels remain stable across sessions, we compute a representational similarity score. For each session we first construct the $C \times C$ Pearson correlation matrix $\mathbf{R}$ from the binned spike data, where $C$ is the number of recorded channels and each entry $r_{ij}$ gives the correlation between the spike trains of channels $i$ and $j$ over the full recording. We then extract the upper-triangular entries (excluding the diagonal) into a vector $\mathbf{u} \in \mathbb{R}^{C(C-1)/2}$ to represent each session. The representational similarity between the source session and a target session is simply the Pearson correlation between $\mathbf{u}^{(0)}$ and $\mathbf{u}^{(k)}$:
\begin{equation}
    S_{\text{rep}} = \text{corr}(\mathbf{u}^{(0)},\; \mathbf{u}^{(k)}).
\end{equation}
A value close to 1 indicates that the channels that were correlated (or uncorrelated) on day 0 remain so on day k. Therefore, the population's internal relational structure is intact. A substantially lower value suggests that the functional relationships among channels have reorganized, which undermines the feature representations learned by the source decoder.

To go beyond pairwise statistics and ask whether the population-level manifold has rotated, we turn to subspace alignment via the Grassmann distance. Raw spike trains are binned into non-overlapping 1-second windows to form a spike-count matrix $\mathbf{X} \in \mathbb{R}^{N \times C}$, where $N$ is the number of bins and $C$ is the number of channels. PCA is fit separately to the source session and the target session, and the top $K=10$ principal components are retained, giving orthonormal basis matrices $\mathbf{U}_0, \mathbf{U}_k \in \mathbb{R}^{C \times K}$ for the corresponding subspaces $\mathcal{U}_0$ and $\mathcal{U}_k$. The principal angles $\theta_1,\dots,\theta_K$ between the two subspaces are computed from the singular values of $\mathbf{U}_0^\top \mathbf{U}_k$. The Grassmann distance is
\begin{equation}
    d_G(\mathcal{U}_0, \mathcal{U}_k) = \sqrt{\sum_{i=1}^{K} \theta_i^2}.
\end{equation}
This quantity is zero when the two subspaces are identical and increases as they rotate apart. A small Grassmann distance means that neural activity on day k still lives in roughly the same low-dimensional manifold as day 0, so the decoder's learned input-output mapping remains geometrically valid. A large distance signals that the dominant directions of variability have shifted, and the source decoder is effectively projecting onto the wrong axes. We use this metric alongside representational similarity to flag sessions where TTA is unlikely to recover acceptable decoding performance from the source model alone.

\section{Experiments and Results}

\subsection{Non-Human Primate Reaching Dataset and TTA Baseline}

All experiments are conducted on the non-human primate reaching dataset from~\cite{odoherty2017reaching}. This dataset contains simultaneously recorded sensorimotor cortex spike trains and reaching kinematics from monkeys performing self-paced reaching movements. In this study, we used 23 sessions from Monkey 1 (Indy), each recorded with 96 channels, collected between 2016-09-15 and 2017-01-24. Following the NeuroBench~\cite{yik2025neurobench}, we encoded the spike data into 4-ms bins and used velocity labels at the same 4-ms resolution. For each session, we included only movement-active recordings and split the data into 50\% training, 25\% validation, and 25\% testing sets. To train the SNN, we used 50-timestep windows for both day-0 pretraining and TTA. During evaluation, testing was performed in a streaming manner, with the SNN processing the full test set continuously.

We use NoMAD \cite{karpowicz2025stabilizing} as the state-of-the-art TTA baseline in our comparisons. NoMAD builds on LFADS~\cite{pandarinath2018inferring} and stabilizes cross-session decoding by aligning neural activity to a shared latent dynamical manifold through unsupervised distribution alignment and spike-rate reconstruction. The model is based on a variational autoencoder with multiple GRU layers. We implemented NoMAD in PyTorch and validated our reproduction on the original dataset from the paper, obtaining similar pretraining and adaptation results. We then transferred the method to our dataset using the same hyperparameters. Spike recordings are binned at 20 ms, and each training sample contained 30 timebins. The training and validation sets are generated with a sliding window and 120 ms overlap. We did not normalize the spike inputs because it had little effect on performance.

\subsection{Experimental Setup}

\begin{table}[t]
\centering
\begin{threeparttable}
\caption{Comparison of test-time adaptation performance across 11 recording sessions from the same non-human primate over a one-month period (Day~0 to Day~22). Models were supervisedly pretrained on the source session \textit{indy\_20161005\_06} (Day~0). Reported values denote $R^2$ scores as mean $\pm$ standard deviation.}
\label{tab:session_results}
\setlength{\tabcolsep}{10pt}
\renewcommand{\arraystretch}{1.15}
\begin{tabular}{c|cc|cc}
\toprule
\multirow{2}{*}{\textbf{Sessions}} & \textbf{LFADS}$^\dagger$ & \multicolumn{1}{c|}{\textbf{NoMAD}$^\dagger$} & \textbf{SNN}$^*$ & \multicolumn{1}{c}{\textbf{MPA} (ours)} \\
& \cite{karpowicz2025stabilizing} & (TTA)~\cite{karpowicz2025stabilizing} & \cite{hueber2024benchmarking} & (TTA) \\
\midrule
\makecell[c]{Day 0 \\ \textbf{(Source)}} 
  & $0.53 \pm 0.05$ & - & $\mathbf{0.59 \pm 0.01}$ & - \\
Day 1  & $-0.25 \pm 0.10$ & $0.33 \pm 0.07$ & $0.10 \pm 0.05$ & $\mathbf{0.36 \pm 0.03}$ \\
Day 2  & $0.21 \pm 0.30$ & $\mathbf{0.46 \pm 0.03}$ & $0.20 \pm 0.03$ & $0.43 \pm 0.04$ \\
Day 6  & $-0.12 \pm 0.19$ & $0.18 \pm 0.08$ & $0.27 \pm 0.05$ & $\mathbf{0.37 \pm 0.05}$ \\
Day 8  & $-0.61 \pm 0.15$ & $-0.57 \pm 0.21$ & $-0.03 \pm 0.01$ & $-0.04 \pm 0.02$ \\
Day 9  & $-0.10 \pm 0.23$ & $\mathbf{0.32 \pm 0.09}$ & $0.19 \pm 0.04$ & $0.28 \pm 0.07$ \\
Day 12  & $0.03 \pm 0.23$ & $\mathbf{0.49 \pm 0.08}$ & $0.11 \pm 0.02$ & $0.46 \pm 0.01$ \\
Day 19  & $-0.21 \pm 0.27$ & $\mathbf{0.37 \pm 0.10}$ & $-0.26 \pm 0.17$ & $0.32 \pm 0.02$ \\
Day 20  & $0.07 \pm 0.20$ & $\mathbf{0.39 \pm 0.06}$ & $0.20 \pm 0.03$ & $0.38 \pm 0.01$ \\
Day 21  & $-0.19 \pm 0.24$ & $\mathbf{0.40 \pm 0.06}$ & $0.03 \pm 0.04$ & $0.37 \pm 0.02$ \\
Day 22 & $-0.20 \pm 0.17$ & $0.28 \pm 0.08$ & $0.12 \pm 0.03$ & $\mathbf{0.40 \pm 0.03}$ \\
\bottomrule
\end{tabular}
\begin{tablenotes}[flushleft]
\footnotesize
\item[$\dagger$] LFADS and NoMAD denote our own implementations in PyTorch.
\item[$*$] SNN denotes our own implementation using soft-reset LIF neurons.
\end{tablenotes}
\end{threeparttable}
\end{table}

For all experiments, we trained three-layer fully connected SNNs. We adapted the training code from~\cite{hueber2024benchmarking} for pretraining. Each hidden layer consisted of 64 LIF neurons with membrane decay $\beta = 0.96$, threshold $V_{\mathrm{th}} = 1.0$, and a fast-sigmoid surrogate gradient with slope 20. Dropout with $p=0.2$ was applied after each linear operation. All models were trained in PyTorch using AdamW. For day-0 pretraining, we used a learning rate of 0.001 and a batch size of 256. Training was run for 10 epochs, and the model with the best validation decoding loss was selected. We used a temporally weighted MSE loss, where the weights increase linearly across the 50-timestep window to place more emphasis on later timesteps. For TTA, we used a learning rate of 0.0005, a batch size of 1024, and a cosine annealing learning-rate scheduler, to train LoRA weights of $r=8$ for the first SNN layer. Training was also run for 10 epochs, and the model with the lowest validation KL-divergence loss was selected. To account for training variability, day-0 pretraining was repeated with five random seeds, and TTA was performed across sessions using the pretrained model from each seed.

\subsection{Month-long adaptation of SNN TTA using MPA}

\begin{figure}[t]
    \centering
    \includegraphics[width=1\linewidth]{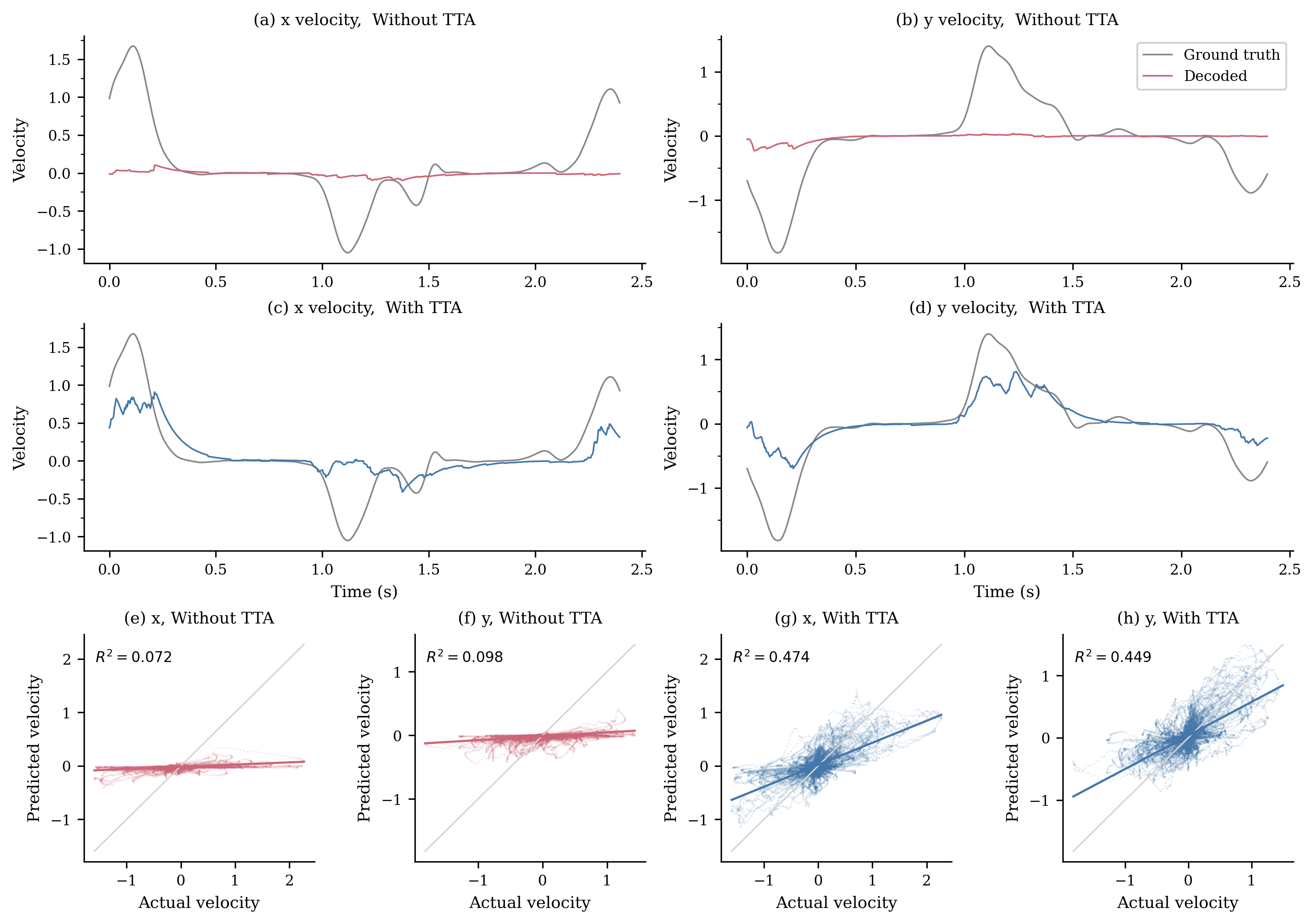}
    \caption{Decoded velocities on session \textit{indy\_20161017\_02} using an SNN pretrained on \textit{indy\_20161005\_06}, with and without MPA. (a–d) A representative 2.4 s segment of the testset selected to highlight the improvement from MPA TTA. Ground truth is shown in gray; decoded velocity in rose (without TTA) or blue (with TTA). (e–h) Predicted vs.\ actual velocity across all test samples. Each panel shows individual samples (scatter), the identity line (light gray), and an ordinary least-squares fit (colored line).}
    \label{fig:decoding}
\end{figure}

\begin{figure}[t]
    \centering
    \includegraphics[width=1\linewidth]{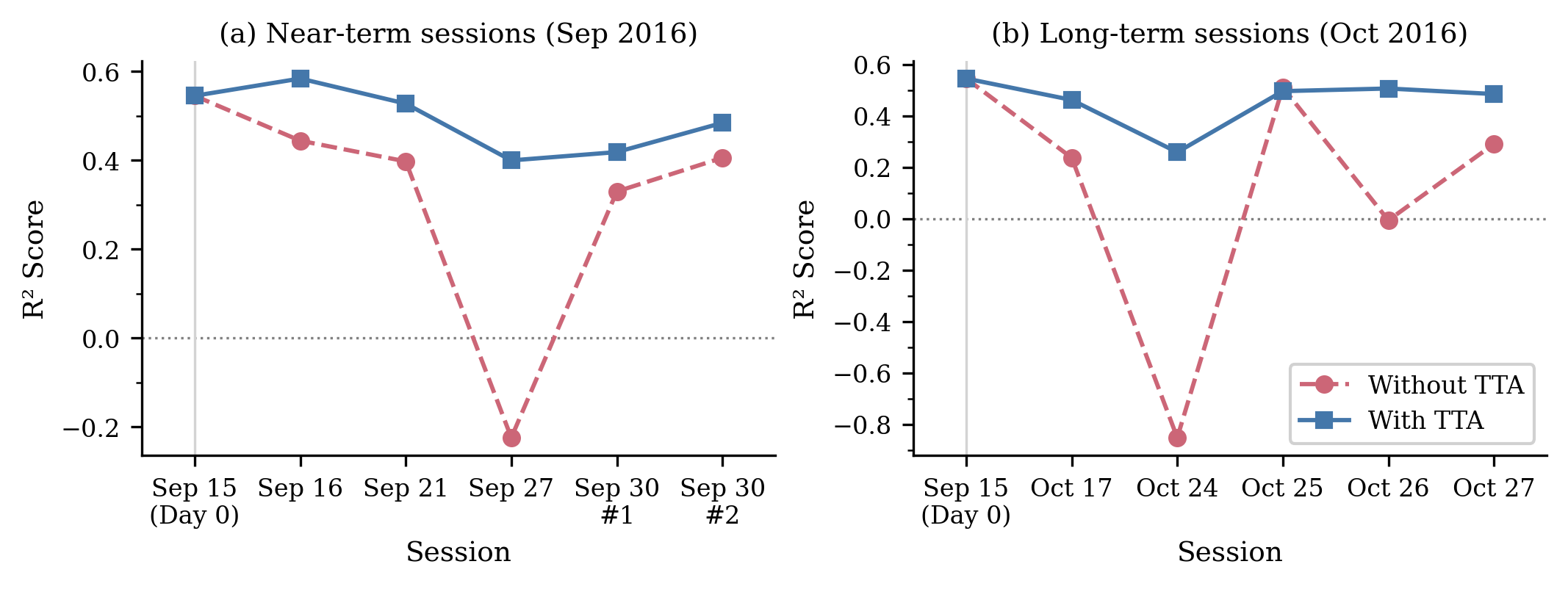}
    \caption{Decoding performance across sessions for an SNN trained on a different Day 0 session (\textit{indy\_20160915\_01}). (a) Near-term adaptation with MPA to five sessions in September 2016. (b) Long-term adaptation with MPA to five sessions in October 2016.}
    \label{fig:sep_day0_tta}
\end{figure}

To evaluate long-term adaptation, we used the first October 2016 session as Day 0 for supervised pretraining and applied TTA to the remaining October sessions. As shown in Table~\ref{tab:session_results}, MPA achieves strong performance throughout this evaluation period. Compared with the non-adaptive SNN baseline, MPA improves performance on 9 of the 10 target sessions and increases the mean $R^2$ from 0.09 to 0.33. MPA is also competitive with the state-of-the-art NoMAD method. It outperforms NoMAD on three positively decoded sessions (Days 1, 6, and 22) and remains close on most of the others. The lower LFADS and NoMAD performance on our benchmark, compared with the results reported in the original paper, likely reflects the greater difficulty of our dataset. Unlike the structured 2D center-out task used in the original paper, our dataset involves self-paced grid reaching with random targets, which is less constrained and generally more challenging. In addition, MPA and the SNN baseline operate at 4 ms resolution, whereas LFADS and NoMAD use 20 ms bins. Both MPA and NoMAD fail on Day 8, indicating a session shift for which unsupervised alignment alone is insufficient. We analyze this failure case further in Section~\ref{sec:fail}.

To highlight MPA's decoding gains, Figure~\ref{fig:decoding} compares SNN ($seed=1$) behavior predictions for the October 17 target session (Day 12) with and without MPA. MPA improves performance from $R^2 \approx 0.08$ to $R^2 \approx 0.46$ for both x- and y-velocity, showing that unsupervised KL-divergence alignment can compensate for cross-session neural drift without movement labels from the new session.

We further provide a complementary view of the TTA behavior when the SNN is pretrained on a different Day 0 session, shown in Fig.~\ref{fig:sep_day0_tta}. Without adaptation, decoding performance degrades markedly as the recording session changes, and in some cases even drops below zero, indicating that the pretrained model no longer generalizes reliably under session drift. In contrast, applying TTA with MPA consistently restores performance for both near-term and long-term sessions, showing that the unsupervised KL-divergence alignment objective can effectively compensate for cross-session neural drift without requiring movement labels from the new session.

Notably, the performance of the non-adapted LFADS and SNN does not show a monotonic decrease with increasing session interval. The absence of a monotonic performance drop with increasing session interval suggests that elapsed time alone is not the main driver of decoding difficulty. Instead, performance is likely dominated by session-specific instabilities in the recorded signals~\cite{dunlap2020classifying,perge2013intra,salatino2017glial}.

\subsection{Target Session Compatibility to Day-0 Session}
\label{sec:fail}

\begin{figure}[t]
    \centering
    \includegraphics[width=1\linewidth]{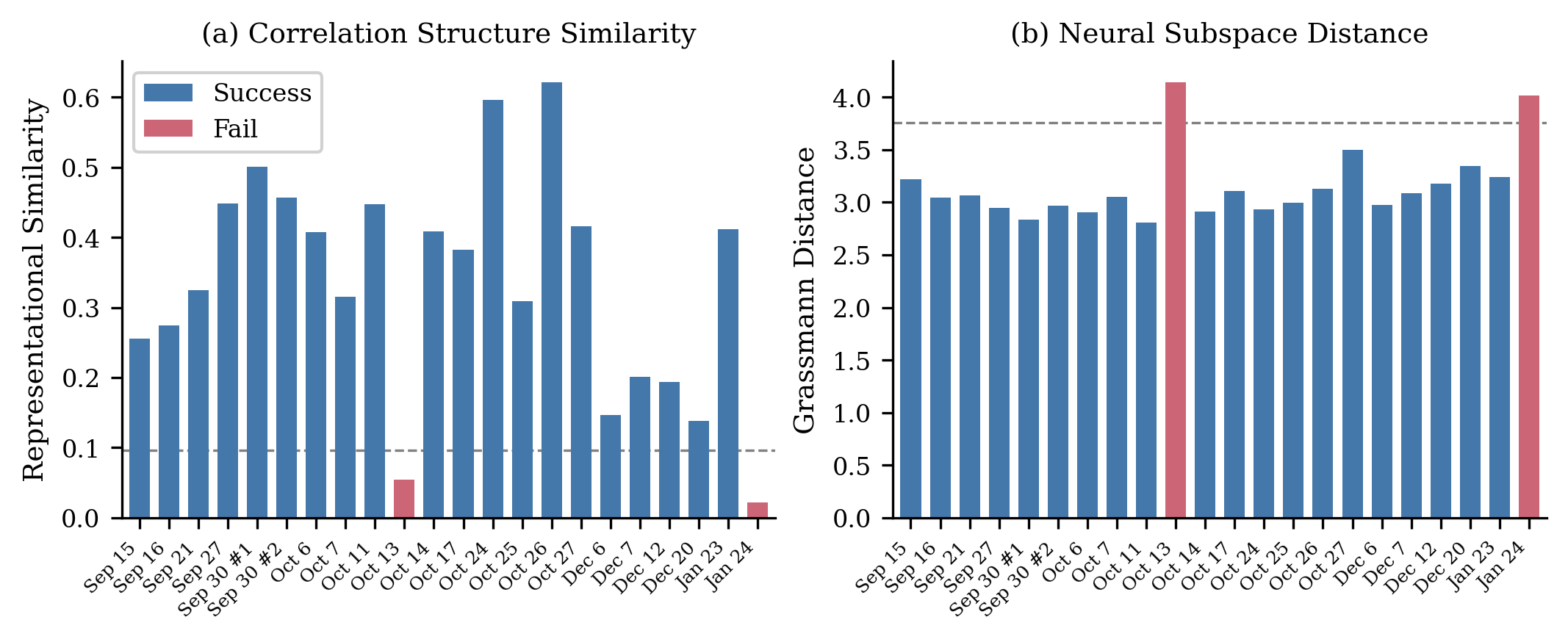}
    \caption{Source–target session compatibility between \textit{indy\_20161005\_06} source session and all 22 target sessions measured by two metrics. (a) Representational similarity for correlation structure similarity. (b) Grassmann distance for neural subspace. Dashed lines indicate empirical thresholds separating sessions where TTA succeeded (blue) from those where it failed (red).}
    \label{fig:fail_session}
\end{figure}

Although our TTA method is effective for most target sessions, Table~\ref{tab:session_results} shows that both NoMAD and MPA fail on the Day 8 session. We therefore examined source-target compatibility between the day-0 source session and this failed session, and asked whether our compatibility metrics could identify other incompatible sessions. To do so, we computed representational similarity and Grassmann distance between the October 5 day-0 source session and the other 22 sessions studied. As shown in Figure~\ref{fig:fail_session}, the failed session on October 13 (Day 8) has the second-lowest representational similarity and the highest Grassmann distance, indicating substantial mismatch in both channel-level structure and low-dimensional subspace. This suggests that TTA fails because the source session is too poorly matched to support reliable unsupervised alignment.

We further examined sessions with low representational similarity and high Grassmann distance. The January 24, 2017 session showed the lowest representational similarity and one of the highest Grassmann distances. For this session, both NoMAD and MPA failed to recover positive decoding performance. In contrast, both methods still worked on December 6 and December 20, despite their relatively low representational similarity. This suggests that TTA fails only when the mismatch is large in both metrics. These metrics may therefore help identify incompatible target sessions and choose the most suitable source session.

\subsection{Effectiveness of Small TTA Training Data}

\begin{figure}[t]
    \centering
    \includegraphics[width=1\linewidth]{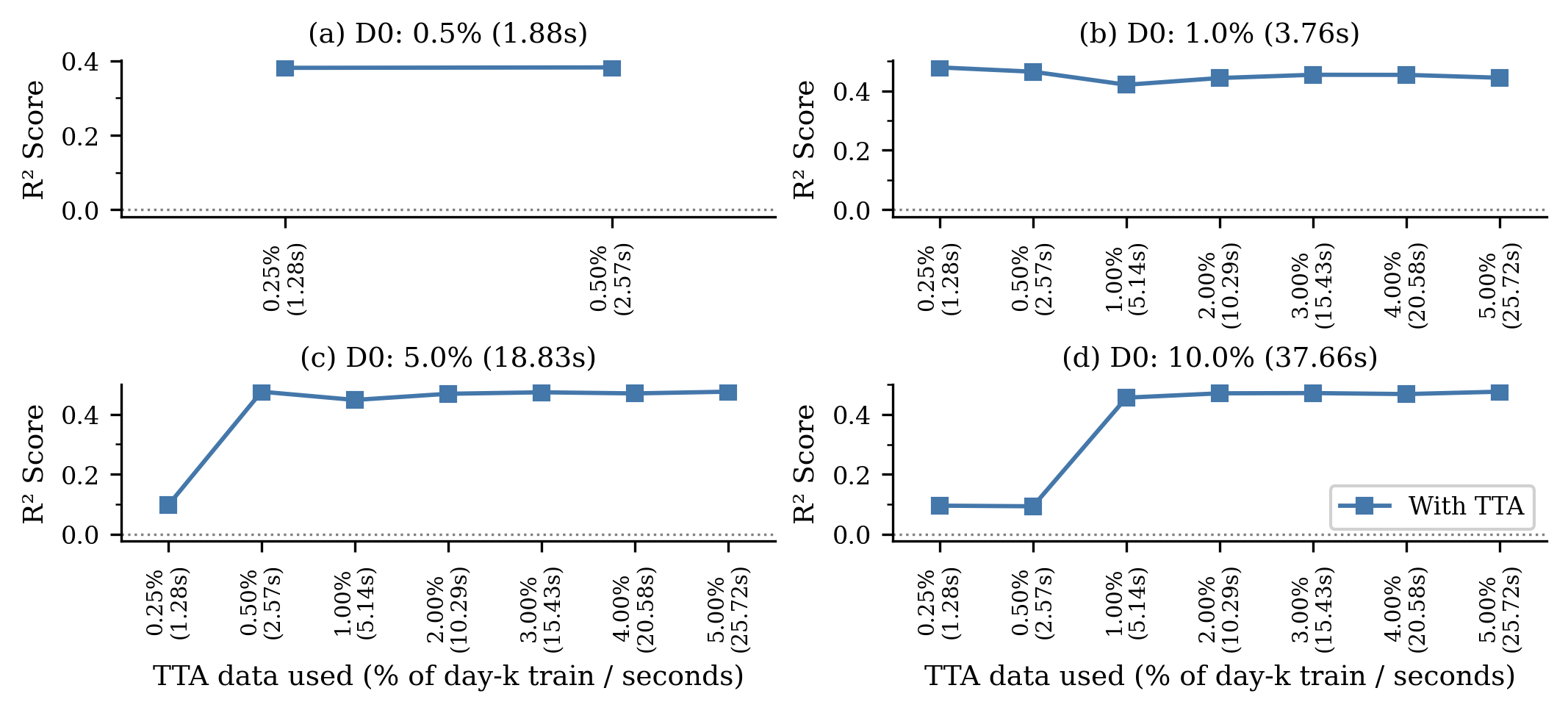}
    \caption{TTA $R^2$ score on the target session (\textit{indy\_20161017\_02}) as a function of TTA alignment data size, for four source day-0 training data sizes. Each panel shows the adapted model $R^2$ after TTA (blue) across seven alignment data fractions (0.25\%–5\% of the target session training set, 1.28–25.72 s). Panel titles indicate the source session data fraction and duration used for TTA alignment.}
    \label{fig:small_data}
\end{figure}

MPA requires both source and target session data for alignment, so in practice, we aim to use as little training data as possible. Smaller datasets reduce memory demands, support edge-device deployment, and enable faster adaptation. To evaluate the data requirements of MPA, we tested different source--target data combinations, with results shown in Figure~\ref{fig:small_data}. TTA still yields a substantial $R^2$ improvement even with very limited data, using as little as 1.88 s of source data and 1.28 s of alignment data. In these experiments, MPA used smaller batch sizes (128 in Figure~\ref{fig:small_data}(a) and 256 in Figure~\ref{fig:small_data}(b-d)), and the number of epochs was adjusted to keep the total number of gradient updates constant.

\section{Conclusion}

We presented MPA, a test-time adaptation method that enables pretrained SNNs to maintain decoding performance across weeks of neural recording drift without behavioral labels. On a month-long non-human primate reaching task, effective adaptation is possible with just seconds of unlabeled data, and the proposed compatibility metrics help identify when adaptation is likely to fail. The lightweight nature of LoRA-based SNN learning makes MPA well-suited for deployment on neuromorphic hardware, where on-device adaptation could eliminate the need for external computing. Performance may be further improved by large-scale pretraining across multiple subjects and sessions to learn more robust initial representations. Extending MPA to continual adaptation and validating it in additional iBCI tasks remain important future directions.

% \subsubsection{\ackname} This publication is part of the project Brain-inspired MatMul-free Deep Learning for Sustainable AI on Neuromorphic Processor with file number NGF.1609.243.044 of the research programme AiNed XS Europe which is (partly) financed by the Dutch Research Council (NWO) under the grant https://doi.org/10.61686/MYMVX53467.

\bibliographystyle{splncs04}
\bibliography{references}

\end{document}